\documentclass[conference]{IEEEtran}
\IEEEoverridecommandlockouts
\usepackage{cite}
\usepackage{amsmath,amssymb,amsfonts}
\usepackage{algorithmic}
\usepackage{graphicx}
\usepackage{textcomp}
\usepackage{xcolor}
\usepackage{verbatim} 
\usepackage{multirow}
\usepackage[inline]{enumitem}
\usepackage{tabularx}
\usepackage{subcaption}
\usepackage[labelformat=parens,labelsep=quad,skip=3pt]{caption}

\def\BibTeX{{\rm B\kern-.05em{\sc i\kern-.025em b}\kern-.08em
		T\kern-.1667em\lower.7ex\hbox{E}\kern-.125emX}}
\begin{document}
	
	\title{BWCFace: Open-set Face Recognition using Body-worn Camera
	}
	
\author{\IEEEauthorblockN{Ali Almadan, Anoop Krishnan, Ajita Rattani}
\IEEEauthorblockA{\textit{Dept. of Electrical Eng. and Computer Science} \\
\textit{Wichita State University, 
Wichita, USA}\\
aaalmadan@shockers.wichita.edu;~ajita.rattani@wichita.edu}
	}
	
	\maketitle
	
	\begin{abstract}
	 With computer vision reaching an inflection point in the past decade, face recognition technology has become pervasive in policing, intelligence gathering, and consumer applications. Recently, face recognition technology has been deployed on body-worn cameras to keep officers safe, enabling situational awareness and providing evidence for trial. However, limited academic research has been conducted on this topic using traditional techniques on datasets with small sample size. This paper aims to bridge the gap in the state-of-the-art face recognition using body-worn cameras (BWC). To this aim, the contribution of this work is two-fold: (1) collection of a dataset called BWCFace consisting of a total of $178$K facial images of $132$ subjects captured using the body-worn camera in in-door and daylight conditions, and (2) open-set evaluation of the latest deep-learning-based Convolutional Neural Network~(CNN) architectures combined with five different loss functions for face identification, on the collected dataset. Experimental results on our BWCFace dataset suggest a maximum of $\textbf{33.89\%}$ Rank-1 accuracy obtained when facial features are extracted using SENet-50 trained on a large scale VGGFace2 facial image dataset. However, performance improved up to a maximum of $\textbf{99.00\%}$ Rank-1 accuracy when pretrained CNN models are fine-tuned on a subset of identities in our BWCFace dataset. Equivalent performances were obtained across body-worn camera sensor models used in existing face datasets. The collected BWCFace dataset and the pretrained/ fine-tuned algorithms are publicly available to promote further research and development in this area. A downloadable link of this dataset and the algorithms is available by contacting the authors.
	\end{abstract}
	
	
	\section{Introduction}
    
	A facial recognition is a technology capable of verifying or identifying a person from a digital image or a video frame from a video source. A typical face recognition pipeline consists of face image acquisition, face detection, facial image representation and matching~\cite{jain2011handbook,ahonen_face_2006}. Face recognition can be categorized as face verification and identification. The former determines whether a pair of face images belong to the same identity, while the latter classifies a face to a specific identity. This is done by comparing the \emph{probe} (test image) to the \emph{gallery} (template) set of all the identities in the dataset. Face recognition is a widely adopted technology in surveillance, border control, healthcare, banking services, and lately, in mobile user authentication with Apple introducing “Face ID” moniker with iPhone X~\cite{jain2011handbook,selfie_bio}. 
	 
	 \begin{figure}[!t]
	\centering
	\includegraphics[scale=0.5]{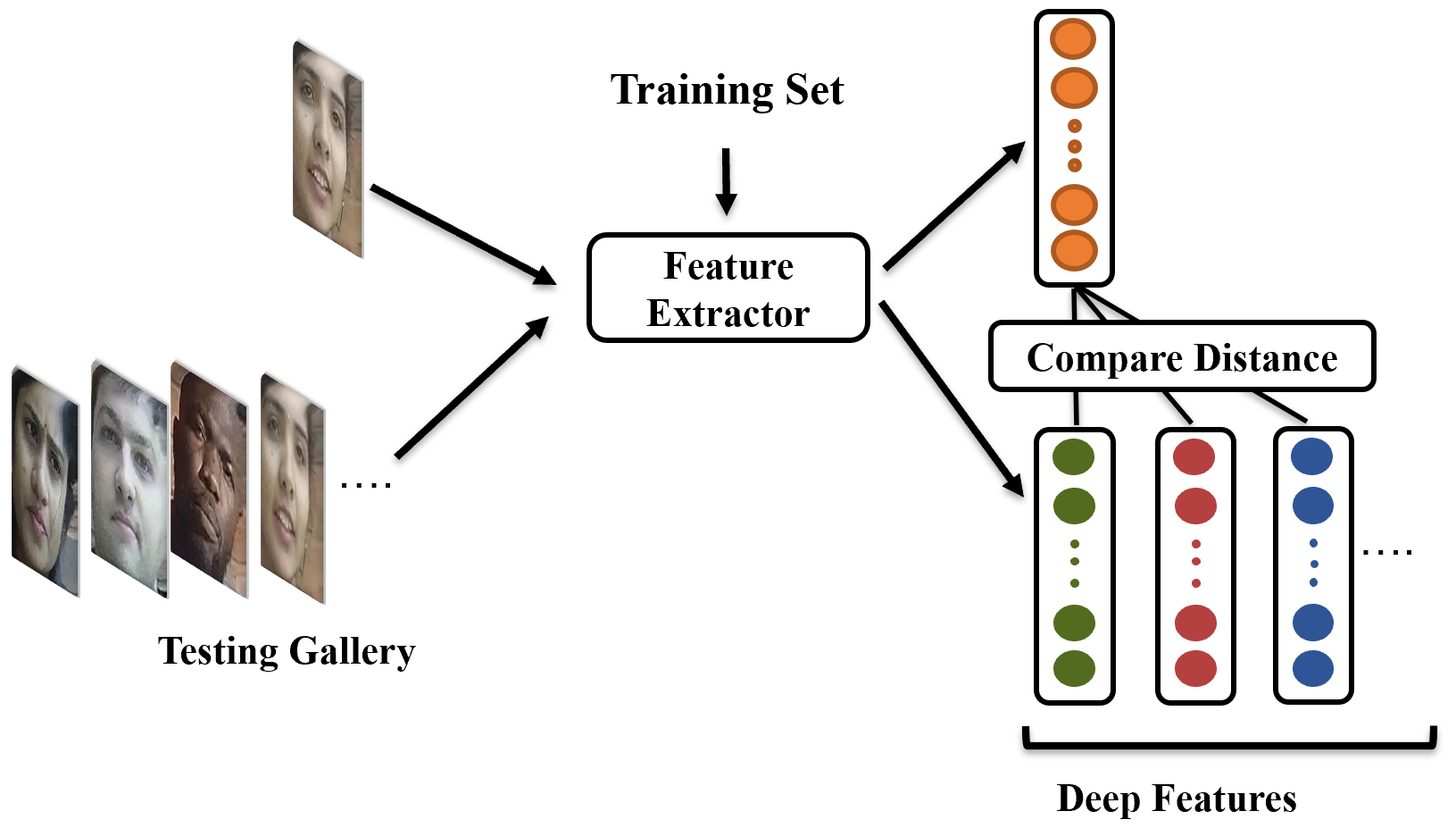}
	\caption{Schema of the open-set face recognition where features are extracted from gallery set and probe via pretrained CNN model. The matching score between the extracted feature vectors is computed using a distance/ similarity metrics for the identity assignment.}
	\label{openset}
\end{figure}

	Over the course of time, various facial image representation methods have been proposed ranging from holistic, such as Principal Component Analysis~(PCA), to local image descriptors, such as Local Binary Patterns~(LBP) and Scale Invariant Feature Transform~(SIFT)~\cite{ahonen_face_2006, bicego_use_2006,kisku}. Recent years have witnessed the great success of convolutional neural networks (CNNs) in face recognition~\cite{minaee2019biometric}. Owing to advanced network architectures of CNNs and discriminative learning approaches with the introduction of the loss functions, face recognition performance has been boosted to an unprecedented level. Loss functions such as ArcFace~\cite{arccite}, CosFace~\cite{coscite}, SphereFace~\cite{spherecite},~and AdaCos~\cite{adacite} have been introduced to learn better feature representation with small intra-class and large inter-class distance for enhancement of the overall accuracy.
	
	Recent interest has been drawn on deploying face recognition technology on \emph{body-worn cameras} (BWC) by law enforcement practitioners in order to keep officers safe, enabling situational awareness and providing evidence for trial~\cite{stanley2015police,blount2015body,bromberg2020public}. A $2012$ report by the National Institute of Justice, within the U.S. Department of Justice, defines body-worn cameras as ``mobile audio and video capture devices that allow officers to record what they see and hear. Devices can be attached to various body areas, including the head or to the body by pocket, and they have the capability to record officer interactions that previously could only be captured by in-car or interrogation room camera systems''~\cite{blount2015body}.
    The widespread use of BWCs has the potential of protecting the public against police misconduct, capturing spontaneous events, subject re-identification, crime scenes investigation or forensic purposes, detecting triggering behaviors, and at the same time, protecting police against a false accusation of abuse~\cite{bromberg2020public}. 
	
	However, only a handful of research has been conducted on face recognition using body-worn cameras~\cite{Al-Obaydy11,brown2016enhanced,bryan2020effects}. These studies have evaluated traditional face recognition methods, such as PCA and Linear Discriminant Analysis (LDA)~\cite{Al-Obaydy11} or face detection methods on the videos captured from the body-worn cameras. The size of the dataset used for evaluation is as small as $20$ subjects~\cite{Al-Obaydy11}. Therefore, the statistical validation of the reported results could not be established. The performance of face identification\footnote{Face recognition and identification are used interchangeably in this paper.} using deep learning methods have not been evaluated until now. This is also attributed to the absence of sufficient sample size datasets. 
	
	The first study evaluating deep learning-based CNN models, such as VGG-16 and ResNet-50, on in-house dataset consisting of $136$K frames from $102$ subjects captured by BWC, is by the authors~\cite{ours}. 
	However, \emph{closed-set evaluation} was performed in this study~\cite{ours}, where the identities in the training and testing set overlaps, usually resulting in higher accuracy. This is because the system better adapts to the subject-specific peculiarities in the dataset. On the contrary, in \emph{open-set evaluation}, identities between the training and testing set do not overlap. In order to be more relevant to the real-world applications at scale, the system needs to perform well in an open-set evaluation where identities in the test set are disjoint from those in the training set. Figure~\ref{openset} illustrates the schema of the deep-learning-based open-set face recognition where the deep features are extracted from a pretrained CNN model. The extracted features from the gallery and the test image are compared using a distance/ similarity metrics for the final identity assignment.

To further advance the research and development, the contribution of this work is as follows:
	\begin{itemize}
	\item A publicly available BWCFace dataset consisting of a total of $178$K facial images from $132$ subjects captured using body-worn cameras in in-door and daylight conditions, and 
	\item Open-set evaluation of the latest deep-learning-based face recognition algorithms\footnote{Algorithms and models are used interchangeably.} based on ResNet~\cite{he2015resnet } and SNENet~\cite{secite} architectures along with five different loss functions, i.e., Softmax, ArcFace~\cite{deng_arcface:_2019}, CosFace~\cite{Wang}, SphereFace~\cite{spherecite}, and AdaCos~\cite{adacite} for face identification on our collected dataset. Rank-1 to Rank-10 identification accuracy values are used for performance evaluation.
	\end{itemize}
	
    The collected dataset and the evaluated algorithms serve as a baseline for further research and development. 
	The dataset and the pretrained CNN models are available by contacting the authors.
	
	This paper is organized as follows: in section \ref{prior_work} literature review of the existing methods on body-worn camera is performed. Section \ref{data_collect} discusses the complete data collection process. Section \ref{exps} discusses the models used and the experimental protocol. Section \ref{eval} discusses the experiments performed and the reported results.

	

	\begin{table}[]
	\caption{Summary of the existing facial analysis studies using body-worn camera.}
	\begin{center}
	 
\begin{tabular}{lll} \hline
\textbf{Reference} & \textbf{Dataset Size} & \textbf{Method} \\ \hline
Al-Obaydy & 20 subjects  & PCA, LDA    \\
   and Sellahewa~\cite{Al-Obaydy11}                        & x 96 images & and Discrete Wavelet
                                      \\ (UBHSD dataset) & \\ \hline
Brown and Fan~\cite{brown2016enhanced} & $638$ face images    & Viola \& Jones, \\ 
                                       &                  & Agg. Feature,               \\
                                       &                   &  Faster R-CNN  \\ \hline
Bryan~\cite{bryan2020effects} &    $3,600$ face images           & Neurotechnology SDKs 
\\ \hline
                                       
Almadan et al.~\cite{ours}           & $102$ subjects & CNN-based face identification \\
             & x $136$K images   & (\textbf{Closed-set evaluation}) \\ \hline
\textbf{PROPOSED} & $132$ subjects  & CNN-based face identification \\ 
                  & x $178$K               & (\textbf{Open-set evaluation}) \\
                  & (BWCFace dataset) & \\ \hline
\end{tabular}
	\end{center}
	\label{summ_lit}
\end{table}

\section{Prior Work}
\label{prior_work}
	\par One of the earliest studies on face recognition using body-worn cameras can be dated back to $2011$ by Al-Obaydy and Sellahewa~\cite{Al-Obaydy11}. Authors collected videos from $20$ subjects using an iOPTEC-P300 body-worn digital video camera and have evaluated traditional subspace methods, namely, PCA, LDA, and discrete wavelet transforms, for facial image representation. The cosine similarity metric was used to compute scores between feature vectors from the pair of facial images for the final identity assignment. Rank-1 recognition accuracy values in the range [$65.83\%$, $76.04\%$] was obtained. This dataset is referred to as University of Buckingham High and Standard Definition (UBHSD) dataset. 
	
	\par Jason et al.~\cite{corso2016video} and Suss et al.~\cite{suss2018design} discussed that the design issues such as video quality, camera-mounting positions, camera activation, and data transfer, are likely to enhance adoption, usability, and acceptance of BWC. The authors in~\cite{suss2018design} also laid down future research directions such as biometric identification, live streaming of videos, and training of the cops.
 
 	\par Brown and Fan~\cite{brown2016enhanced} evaluated three face detectors: Viola \& Jones, Aggregate Channel Feature, and Faster R-CNN over a collected dataset containing $638$ face images captured from BWC. 
	 R-CNN approach achieved the highest accuracy of $94\%$ over other face detection methods. 
	 The challenges included detecting blurry, dark, and occluded faces.  
	
	 \par  Bryan~\cite{bryan2020effects} studied the impact of the location of the body-worn camera on the facial detection performance. The reported results suggest that the most ideal and preferred location for camera mounting is at a center chest. 

	
	\par In our prior work~\cite{ours}, deep CNN models, i.e., VGG and ResNet, were evaluated for face identification on an in-house dataset consisting of $136$k frames from $102$ subjects. In this study, \emph{closed-set evaluation} was adopted.
	Overall, the mean performance Rank-$1$ accuracy of $99.5\%$ was obtained for all the models in the same lighting conditions, but the accuracy dropped to $98.26\%$ in the cross-lighting conditions. 
	
	Table~\ref{summ_lit} summarizes the existing studies on facial analysis using a body-worn camera. As can be seen, mostly traditional methods and small size datasets are used in existing studies. 
	
	\section{BWCFace Data Collection}
	\label{data_collect}
	
	Students from Wichita State University were recruited for the dataset collection using the Body-worn Camera (BWC). Videos from $132$ subjects were collected from Boblov $1296$P WiFi Body Mounted Camera\footnote{https://www.boblov.com/}. The subjects were recorded using a $30$s high definition video of resolution $1920\times1080$ pixels at $30$fps in in-door and daylight settings ($10$ to $15$ minutes apart). The body-worn camera was mounted on the center chest of the Research Assistant following the recommendation in~\cite{bryan2020effects}. During recording, subjects were asked to act naturally in order to represent realistic conditions. The actions captured from the subjects include talking, facial expressions, and head movements. The recording was done in an uncontrolled environment with non-uniform background and varying distance between the subject and the body-worn camera. However, the distance remained reasonable, subjects collaborated with BWC carrier to represents an active surveillance scenario. Ethnic groups of the captured subjects include Asian, White American, Black, and Middle Eastern.

\begin{figure}[!t]
		\centering
		\includegraphics[scale=2]{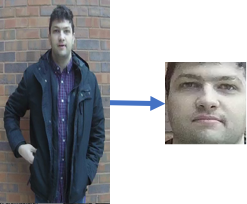}
		\caption{Face detection on the video frame of a randomly chosen subject from BWCFace dataset using Dlib library~\cite{king2009dlib}.}
		\label{fig_det}
	\end{figure}

\begin{figure}[!t]
		\centering
		\includegraphics[scale=0.7]{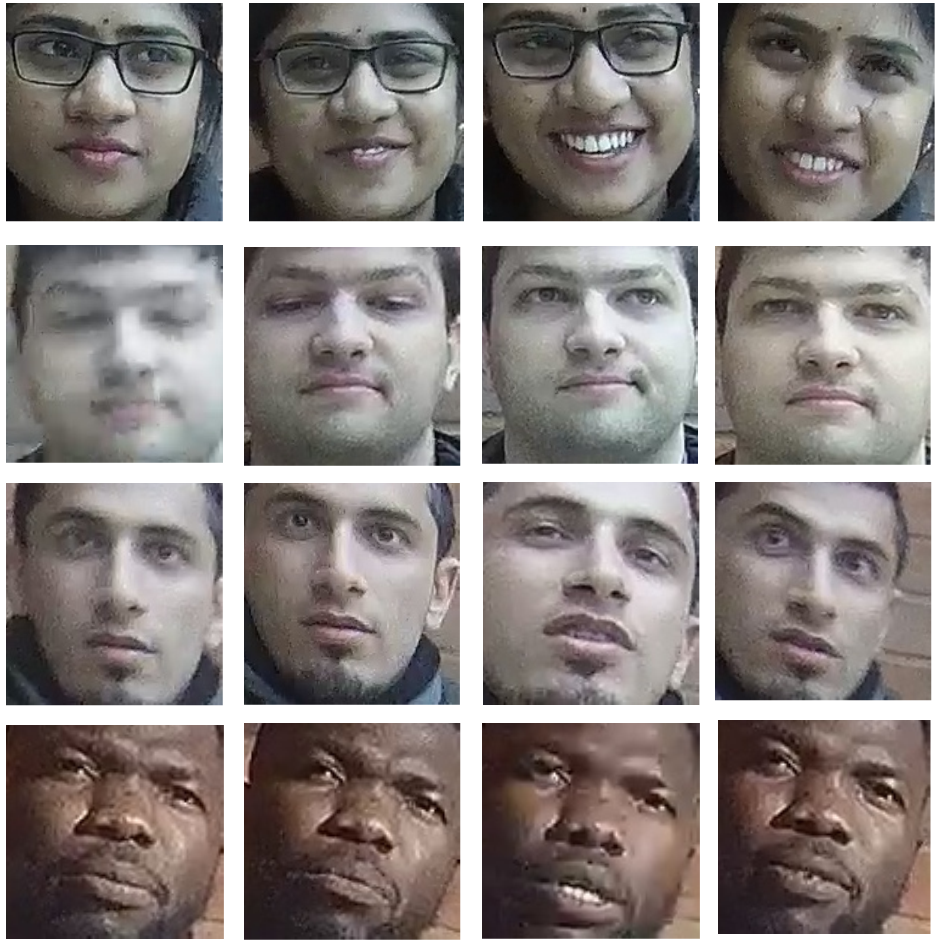}
		\caption{Subset of cropped face images from randomly selected subjects in the BWCFace dataset. Facial expression, motion blur, and poses are clearly visible in the samples.}
		\label{fig_data}
	\end{figure}

	Video clips were processed on a frame-by-frame basis for face detection using Dlib library~\cite{king2009dlib}. The Dlib face detector is based on Histogram of Oriented Gradients (HOG) and Support Vector Machines (SVM).
	False positives in the form of non-face images were manually removed. 
	This results in the collection of a total of $178$K images from $132$ subjects (varying size facial crops) equally divided between in-door and daylight conditions.  Figure~\ref{fig_det} shows the face region cropped from one of the video frames of the captured subject using Dlib library. Figure~\ref{fig_data} shows the subset of cropped face images in the dataset from randomly selected subjects. The intra-class variations such as facial expressions, pose, and motion blur are clearly visible in the images. 

	\section{Implementation details}
	\label{exps}
	In this section, we discuss the CNN architectures used, along with several loss functions and the experimental protocol. All the experiments are conducted on a workstation configured with AMD Ryzen 5 processor and GeForce RTX 2060 GPU. PyTorch\footnote{https://pytorch.org/} framework was used for CNN model implementation and evaluation.

	\subsection{CNN Architectures}
	We used the popular ResNet~\cite{he2015resnet} and SE-ResNet~\cite{secite} architectures as they are widely used for face recognition. 
	These two backbone architectures are described below.

	\begin{itemize}
	\item \textbf{ResNet-50}: ResNet is a short form of residual network based on  the  idea  of  ``identity  shortcut  connection''  where input  features  may  skip  certain  layers~\cite{he2015resnet}.  In this study, we  used  ResNet-$50$ which has $23.5$M parameters.
	
	\item \textbf{SE-ResNet-50}: The added block SE, the Squeeze-and-Excitation, utilizes global information in selecting representative features. In other words, the block selectively adds weights to the feature maps that are representative~\cite{secite}. 
	
	\end{itemize}
	
	Briefly, ResNet-50~\cite{he2015resnet} and SE-ResNet-50~\cite{secite} (SENet) are trained from scratch on VGGFace2~\cite{cao2018vggface2} and MS1Mv2~\cite{guo2016ms} along with five different loss functions i.e., CosFace~\cite{Wang}, ArcFace~\cite{deng_arcface:_2019}, SphereFace~\cite{liu_sphereface:_2017}, AdaCos~\cite{adacite}, and the default Softmax~\cite{deng_arcface:_2019}. These cosine loss functions (ArcFace, CosFace, SphereFace, and AdaCos) aim to maximise face class separablity. Readers are referred to the cited papers on more information on these loss functions.
	
	\subsection{Network Training}
	\label{extract_lbl}
	The CNN models are trained from scratch on MS1Mv2~\cite{deng_arcface:_2019} and VGGFace2~\cite{cao2018vggface2} datasets, separately. We have used the curated version of these datasets for training purposes. For the purpose of experiments, all the images were resized to $224\times 224$. VGGFace2~\cite{cao2018vggface2} dataset contains $3.31$ million images from $9,131$ celebrities. For training the models, we use a training split of $3.14$ million images from $8,631$ subjects.
	The MS1Mv2 dataset~\cite{deng_arcface:_2019}, is a cleaned version of the MS1M dataset~\cite{guo2016ms}, containing around $5.8$ million images from $85,742$ subjects were used for training the models. 

	For ResNet-$50$ model based on cosine loss functions (ArcFace, CosFace, SphereFace, and AdaCos), Batch-normalization, Drop-out, 
	and fully connected layers of $2048$ and $512$ were added after the last convolutional layer. This was followed by the last fully connected layer of $512$ and the final output layer equal to the number of classes for the network training. The angular margin for ArcFace, CosFace, and SphereFace  was set to $0.50$, $0.40$ and $4.0$ following the default setting reported in their respective papers. AdaCos automatically adjusts the scale parameter. The networks were trained using an Adam optimizer~\cite{adamcite} with a batch size of $128$ for $25$ epochs using an early stopping mechanism on the validation set.	The learning rate was set equal to 1e-4 and a decay of 5e-4 . 
	
	Table~\ref{tune-acc} shows the validation accuracy obtained on training the network on two datasets along with five different loss functions. The validation accuracy in the range of [$94.63\%$, $99.85\%$] was obtained. Mostly, AdaCos and Softmax obtained equivalent validation accuracy values.

	

	

	\begin{table} [h]
		\centering
		\caption{Validation accuracy for CNN models trained on MS1Mv2~\cite{deng_arcface:_2019} and VGGFace2~\cite{cao2018vggface2} datasets.}
		\begin{tabular}{|c|c|c|c|} 
			\hline
			\textbf{CNN}         & \textbf{Trained}                  & \textbf{Loss Function} & \textbf{Accuracy [\%]}  \\ 
			\hline
			\multirow{5}{*}{ResNet-50}     & \multirow{5}{*}{MS1Mv2} & CosFace       & 97.88                \\
			&                            & ArcFace       & 97.67                \\
			&                            & SphereFace    & 99.04                \\
			&                            & AdaCos        & \textbf{99.75}                \\
			&                            & Softmax        & 99.58                \\ 
			\hline
			\multirow{5}{*}{ResNet-50} & \multirow{5}{*}{VGGFace2} & CosFace       & 97.83                \\
			&                            & ArcFace       & 97.58                \\
			&                            & SphereFace    & 99.13                \\
			&                            & AdaCos        & 99.58                \\
			&                            & Softmax        & \textbf{99.63}                \\ 
			\hline
			\multirow{5}{*}{SENet-50}     & \multirow{5}{*}{MS1Mv2}  & CosFace       & 95.29                \\
			&                            & ArcFace       & 94.63                \\
			&                            & SphereFace    & 97.54                \\
			&                            & AdaCos        & 98.92                \\
			&                            & Softmax        & \textbf{99.21}                \\ 
			\hline
			\multirow{5}{*}{SENet-50} & \multirow{5}{*}{VGGFace2}  & CosFace       & 96.96                \\
			&                            & ArcFace       & 97.08                \\
			&                            & SphereFace    & 98.67                \\
			&                            & AdaCos        & 99.50                \\
			&                            & Softmax        & \textbf{99.85}                \\
			\hline
		\end{tabular}
		\label{tune-acc}
	\end{table}

	



\subsection{Experimental Protocol}
For the open-set evaluation of the trained deep learning models for face identification on BWCFace dataset, the gallery set consists of $12$ face images per subject. The probe set consists of $100$ face images randomly selected per subject. 

From the gallery and probe set of all the subjects, deep features are extracted using the pretrained ResNet-50 and SENet-50 models (discussed in Section IV).
The matching score between the deep features from the pair of gallery and probe image is computed using cosine similarity, as given below: 
	

	\[
	\operatorname{sim}(u, v)
	= \frac{u \cdot v}{|u||v|}
	= \frac{\sum_{i = 1}^N u_i v_i}
	{\sqrt{\left(\sum_{i = 1}^N u_i^2\right)
			\left(\sum_{i = 1}^N v_i^2\right)}}
	\]

	where u and v are the two deep feature vectors: $u = \{u_1,u_2,\dots,u_N\}$ 
	and $v = \{v_1,v_2,\dots,v_N\}$.

The average of the scores from the multiple gallery images per subject is used for final identity assignment. In cross-lighting conditions, the template (gallery set) was selected from a different lighting condition that of probe set for all the identities.

	\begin{table}[]
		\caption{Open-set face identification accuracy of the models on BWCFace dataset when trained on MS1Mv2~\cite{deng_arcface:_2019} and VGGFace2~\cite{cao2018vggface2} datasets. Comparative evaluation is done on UBHSD dataset~\cite{Al-Obaydy11}.}
		\begin{center}

		\begin{tabular}{|c|c|c|c|}
			\hline
			\textbf{Light Condition - Dataset} & \multicolumn{1}{c|}{\textbf{Rank-1 [\%]}} & \multicolumn{1}{c|}{\textbf{Rank-5 [\%]}} & \textbf{Rank-10 [\%]} \\ \hline
			\multicolumn{4}{|c|}{\textbf{ResNet-50  -  MS1Mv2}}                                                                                      \\ \hline
			Office vs. Office            & 28.18                                 & 51.87                                 & 63.56             \\ \cline{1-1}
			Day. vs. Day                 & 30.47                                 & 55.78                                 & 70.58             \\ \cline{1-1}
			Day vs. Office               & 2.31                                  & 10.96                                 & 21.33             \\ \cline{1-1}
			Office vs. Day               & 2.14                                  & 10.87                                 & 22.33             \\ \cline{1-1}
			\textbf{UBHSD~\cite{Al-Obaydy11}}                              & 20.89                                 & 47.62                                 & 68.10             \\ \hline
			\multicolumn{4}{|c|}{\textbf{ResNet-50  -  VGGFace2}}                                                                                  \\ \hline
		 Office vs. Office            & 26.43                                 & 46.53                                 & 56.93             \\ \cline{1-1}
			Day. vs. Day                 & 26.55                                 & 46.35                                 & 60.45             \\ \cline{1-1}
			 Day vs. Office               & 3.94                                  & 14.57                                 & 24.24             \\ \cline{1-1}
		Office vs. Day               & 3.09                                  & 15.49                                 & 29.35             \\ \cline{1-1}
			\textbf{UBHSD~\cite{Al-Obaydy11}}                              & 21.85                                 & 54.82                                 & 73.63             \\ \hline
			\multicolumn{4}{|c|}{\textbf{SENet-50  -  MS1Mv2}}                                                                                       \\ \hline
		Office vs. Office            & 30.02                                 & 56.09                                 & 68.57             \\ \cline{1-1}
			Day. vs. Day                 & 29.27                                 & 55.09                                 & 69.40             \\ \cline{1-1}
			Day vs. Office               & 2.45                                  & 10.10                                 & 20.24             \\ \cline{1-1}
			Office vs. Day               & 1.77                                  & 10.39                                 & 22.38             \\ \cline{1-1}
			\textbf{UBHSD~\cite{Al-Obaydy11}}                              & 21.07                                 & 47.74                                 & 68.63             \\ \hline
			\multicolumn{4}{|c|}{\textbf{SENet-50  -  VGGFace2}}                                                                                   \\ \hline
		Office vs. Office            & 28.64                                 & 50.73                                 & 61.77             \\ \cline{1-1}
		Day. vs. Day                 & \textbf{33.89}                                 & \textbf{61.69}                                 & \textbf{75.06}            \\ \cline{1-1}
			Day vs. Office               & 2.71                                  & 12.69                                 & 22.76             \\ \cline{1-1}
		Office vs. Day               & 2.78                                  & 15.12                                 & 25.74             \\ \cline{1-1}
			\textbf{UBHSD~\cite{Al-Obaydy11}}                              & 29.94                                 & 58.87                                 & 73.99             \\ \hline
		\end{tabular}
			\end{center}
		\label{tab:scratch}
	\end{table}

	\section{Evaluation Results}
	\label{eval}
	In this section, we evaluate the open-set performance of the face recognition on our BWCFace dataset collected using a body-worn camera. All the models are evaluated in the same and cross-lighting conditions. Next, we discuss the experiments performed and the results obtained.\newline

 \noindent \textbf{1. Pre-trained CNNs:} 
    First, we evaluated the open-set performance of the face recognition system on BWCFace dataset when deep features are extracted from ResNet-50 and SENet-50, trained on MS1Mv2~\cite{deng_arcface:_2019} and VGGFace2~\cite{cao2018vggface2} datasets, in the same and cross-lighting conditions. 
 
    Table~\ref{tab:scratch} shows the open-set evaluation of these pretrained face recognition models on BWCFace dataset. Rank-$1$, Rank-$5$, and Rank-$10$ accuracy values are recorded. These results are shown only for the Softmax loss function. Equivalent performances were obtained for other loss functions, not shown for the sake of space. Cross-lighting scenario is shown using Day vs. Office entry and vice-versa. Day vs. Office entry means that gallery set is acquired in daylight and probe set in office light conditions.
	
	It can be seen that overall low accuracy values are obtained. SENet-50 has slightly better performance than ResNet-50. Day vs. Day obtained better accuracy values over Office vs. Office. Maximum accuracy of $33.89\%$, $61.69\%$, and $75.06\%$ are consistently obtained at Rank-1, Rank-5, and Rank-10 for SENet-50 when trained on VGGFace2 dataset. When compared with the same lightning settings, cross lighting evaluation degraded the accuracy values by a factor of $5.18$,$2.35$, and $1.75$ at Rank-1, Rank-5, and Rank-10, respectively. We have also reported an open-set evaluation of the pretrained models on the UBHSD dataset~\cite{Al-Obaydy11}. Equivalent accuracy values were obtained when compared to the same lighting condition results on our BWCFace dataset. UBHSD dataset did not annotate the images based on the lighting condition. Therefore, cross-lighting comparisons could not be performed.
	
	\emph{Overall, low accuracy values were obtained when pretrained CNN models are used for face recognition on our BWCFace and UBHSD dataset acquired using the body-worn camera}. This could be due to existing face image datasets scraped from the web are not representative of the images captured using the body-worn camera. \newline

\begin{table}
\centering
\caption{Rank-1, Rank-5 and Rank-10 accuracy values of the CNN models trained on MS1Mv2~\cite{deng_arcface:_2019} and VGGFace2~\cite{cao2018vggface2} datasets and fine-tuned on subset of subjects from BWCFace dataset and evaluated on UBHSD dataset~\cite{Al-Obaydy11}.} \label{ubhsd-results}
\resizebox{\columnwidth}{!}{

\begin{tabular}{|c|c|c|c|c|c|} 
\hline
\textbf{CNN }                  & \textbf{Trained}                   & \textbf{Metrics}    & \textbf{Rank-1 [\%]} & \textbf{Rank-5 [\%]} & \textbf{Rank-10 [\%]}  \\ 
\hline
\multirow{5}{*}{ResNet-50} & \multirow{5}{*}{MS1Mv2}     & CosFace    & 98.21   & 99.52   & 99.94     \\
                           &                           & ArcFace    & 97.92   & 99.35   & 99.88     \\
                           &                           & SphereFace & 98.51   & 99.70   & 99.94     \\
                           &                           & AdaCos     & 97.26   & 99.40   & 99.82     \\
                           &                           & Sofmax     & 98.99   & 99.88   & 100.00    \\ 
\hline
\multirow{5}{*}{ResNet-50} & \multirow{5}{*}{VGGFace2} & CosFace    & 98.39   & 99.70   & 99.94     \\
                           &                           & ArcFace    & 98.15   & 99.64   & 99.82     \\
                           &                           &  SphereFace & 98.27   & 99.64   & 99.76     \\
                           &                           & AdaCos     & 97.92   & 99.52   & 99.64     \\
                           &                           & Sofmax     & 98.75   & 99.64   & 99.76     \\ 
\hline
\multirow{5}{*}{SENet-50}  & \multirow{5}{*}{MS1Mv2}     & CosFace    & 96.07   & 98.21   & 99.11     \\
                           &                           & ArcFace    & 95.60   & 98.33   & 98.93     \\
                           &                           & SphereFace & 96.19   & 98.39   & 98.87     \\
                           &                           & AdaCos     & 95.83   & 98.51   & 99.40     \\
                           &                           & Sofmax     & 95.71   & 98.69   & 99.82     \\ 
\hline
\multirow{5}{*}{SENet-50}  & \multirow{5}{*}{VGGFace2} & CosFace    & 97.02   & 98.57   & 99.35     \\
                           &                           & ArcFace    & 97.08   & 98.63   & 99.52     \\
                           &                           & SphereFace & 97.44   & 98.57   & 99.40     \\
                           &                           & AdaCos     & 96.55   & 98.81   & 99.40     \\
                           &                           & Sofmax     & 97.98   & 99.52   & 99.82     \\
\hline
\end{tabular} }
\label{finetune_SNE1}
\end{table}

\noindent \textbf{2. Pretrained CNNs Fine-tuned on BWCFace Dataset:} 
Table~\ref{finetun_BWC} shows the performance of the ResNet-50 model, pretrained on VGGFace2~\cite{cao2018vggface2} and MS1Mv2~\cite{deng_arcface:_2019}, when fine-tuned on subset of $30$ subjects from our BWCFace dataset. It can be seen that the validation accuracy of $99\%$ was obtained by SphereFace, AdaCos, and Softmax. CosFace and ArcFace obtained around $97\%$ validation accuracy. 

Open set evaluation obtained a maximum of $99\%$ Rank-1 for Softmax loss function in Day vs. Day settings. At Rank-5, all the loss functions obtained around $99\%$ accuracy for Day vs.Day settings. At rank-10, all the loss functions hit $99\%$ accuracy in the same lighting conditions.

By comparing with the mean accuracy over all lighting conditions, the average drop in the performance for cross lighting conditions was $8.97\%$, $2.76\%$, and $1.31\%$ at Rank-1, Rank-5, and Rank-10, respectively.
Overall, the drop in the performance was higher for Day vs. Office in comparison to Office vs. Day condition. 

\par Table~\ref{finetune_SNE1} shows the performance of the SENet-50 when trained on MS1Mv2~\cite{deng_arcface:_2019} and VGGFace2~\cite{cao2018vggface2} and fine-tuned on $30$ subjects from our BWCFace dataset. It can be seen that the validation accuracy of $99\%$ was obtained by AdaCos and Softmax. CosFace and ArcFace obtained around $97\%$ validation accuracy. 
Open set evaluation obtained a maximum of $99\%$ Rank-1 for Softmax loss function in Day vs.Day settings, when pre-trained weights are obtained from the MS1Mv2 dataset. At Rank-5, most of the models obtained around $99\%$ accuracy in day vs. day setting. At Rank-10, most of the models obtained $99\%$ accuracy for the same lighting condition.
The average drop in the performance for cross lighting conditions was $7.23\%$, $2.02\%$ and $0.80\%$ at Rank-1, Rank-5, and Rank-10, respectively.
 The drop in the performance was higher for Day vs. Office in comparison to Office vs. Day condition.

The comparison in terms of architectures showed that SENet-50 obtained slightly better identification performance compared to ResNet-50 for both the datasets. In fact, the average Rank-1 accuracy for ResNet-50 is $88.55\%$, and $89.09\%$ for SENet-50. This can be due to the significant influence of the global information that the SE block adds. \emph{Overall, fine-tuning the models on a small subset of subjects captured from body-worn camera has obtained hallmark performance.}\newline

	
	
	
\noindent \textbf{3. Fine-tuned Models Evaluated on UBHSD Dataset~\cite{Al-Obaydy11}}. 
	We also evaluated open-set face recognition performance of the models fine-tuned on our BWCFace dataset and tested on the UBHSD dataset~\cite{Al-Obaydy11}. The UBHSD dataset contained $96$ face images from each of the $20$ distinct subjects. This dataset was collected using a iOPTEC-30 BWC camera sensor which is different from our sensor. The dataset comprised of blurred and occluded face images with eyes half-closed. For this particular dataset, templates (gallery set) consist of $12$ face images and probe set consist of $84$ images. The deep features from the pair of template and probe images are extracted using the fine-tuned models. As this dataset is not annotated with the lighting condition, cross-lighting evaluation could not be performed.
	Table~\ref{ubhsd-results} shows the Rank-1,  Rank-5  and  Rank-10  accuracy  values of the CNN models fine-tuned on our BWCFace dataset and tested on UBHSD dataset. It can be seen that maximum Rank-1 accuracy of around $98\%$ (1\% performance drop) for most of the loss functions. At Rank-5 and Rank-10, most of the loss functions obtained $99\%$ accuracy. At Rank-1 and Rank-5, ResNet-50 obtained better performance than SENet-50. ResNet-50 obtained a mean of $98.24\%$ and $99.60\%$ for Rank-1 and Rank-5, respectively, whereas SENet-50 obtained a mean of $96.55\%$ for Rank-1 and $98.63\%$ for Rank-5, respectively.
	
	
	It was observed that the performance for all models and loss functions was comparable with minimum drop, despite that they were fine-tuned on our dataset (see Table ~\ref{ubhsd-results}). \emph{This also suggest that equivalent performances of fine-tuned models could be obtained across different body-worn camera models}.

	\begin{table*}[h]
		\centering
		\caption{Rank-1, Rank-5 and Rank-10 accuracy values of the ResNet-50 trained on MS1Mv2~\cite{deng_arcface:_2019} and VGGFace2~\cite{cao2018vggface2} datasets, fine-tuned and evaluated on non-overlapping subjects from BWCFace dataset.}
		\begin{tabular}{cllccc}
			\hline
			\multicolumn{1}{|c|}{\textbf{Loss Function}}      & \multicolumn{1}{c|}{\textbf{Validation Accuracy [\%]}} & \multicolumn{1}{c|}{\textbf{Light. Condition}} & \multicolumn{1}{c|}{\textbf{Rank-1 [\%]}} & \multicolumn{1}{c|}{\textbf{Rank-5 [\%]}} & \multicolumn{1}{c|}{\textbf{Rank-10 [\%]}} \\ \hline
			\multicolumn{6}{|c|}{\textbf{ResNet-50  -  MS1Mv2}}                                                                                                                                                                                                                           \\ \hline
			\multicolumn{1}{|c|}{\multirow{4}{*}{CosFace}}    & \multicolumn{1}{c|}{\multirow{4}{*}{97.88}}      & \multicolumn{1}{c|}{Office vs. Office}         & \multicolumn{1}{c|}{94.19}           & \multicolumn{1}{c|}{98.33}           & \multicolumn{1}{c|}{99.20}            \\
			\multicolumn{1}{|c|}{}                            & \multicolumn{1}{c|}{}                            & \multicolumn{1}{c|}{Day. vs. Day}              & \multicolumn{1}{c|}{98.08}           & \multicolumn{1}{c|}{99.46}           & \multicolumn{1}{c|}{99.63}            \\
			\multicolumn{1}{|c|}{}                            & \multicolumn{1}{c|}{}                            & \multicolumn{1}{c|}{Day vs. Office}            & \multicolumn{1}{c|}{76.45}           & \multicolumn{1}{c|}{90.41}           & \multicolumn{1}{c|}{94.27}            \\
			\multicolumn{1}{|c|}{}                            & \multicolumn{1}{c|}{}                            & \multicolumn{1}{c|}{Office vs. Day}            & \multicolumn{1}{c|}{83.83}           & \multicolumn{1}{c|}{93.98}           & \multicolumn{1}{c|}{96.89}            \\ \hline
			\multicolumn{1}{|c|}{\multirow{4}{*}{ArcFace}}    & \multicolumn{1}{c|}{\multirow{4}{*}{97.66}}      & \multicolumn{1}{c|}{Office vs. Office}         & \multicolumn{1}{c|}{93.41}           & \multicolumn{1}{c|}{98.11}           & \multicolumn{1}{c|}{99.04}            \\
			\multicolumn{1}{|c|}{}                            & \multicolumn{1}{c|}{}                            & \multicolumn{1}{c|}{Day. vs. Day}              & \multicolumn{1}{c|}{97.83}           & \multicolumn{1}{c|}{99.38}           & \multicolumn{1}{c|}{99.68}            \\
			\multicolumn{1}{|c|}{}                            & \multicolumn{1}{c|}{}                            & \multicolumn{1}{c|}{Day vs. Office}            & \multicolumn{1}{c|}{75.22}           & \multicolumn{1}{c|}{90.27}           & \multicolumn{1}{c|}{94.25}            \\
			\multicolumn{1}{|c|}{}                            & \multicolumn{1}{c|}{}                            & \multicolumn{1}{c|}{Office vs. Day}            & \multicolumn{1}{c|}{82.55}           & \multicolumn{1}{c|}{93.15}           & \multicolumn{1}{c|}{96.43}            \\ \hline
			\multicolumn{1}{|c|}{\multirow{4}{*}{SphereFace}} & \multicolumn{1}{c|}{\multirow{4}{*}{99.04}}      & \multicolumn{1}{c|}{Office vs. Office}         & \multicolumn{1}{c|}{94.14}           & \multicolumn{1}{c|}{98.28}           & \multicolumn{1}{c|}{99.17}            \\
			\multicolumn{1}{|c|}{}                            & \multicolumn{1}{c|}{}                            & \multicolumn{1}{c|}{Day. vs. Day}              & \multicolumn{1}{c|}{98.34}           & \multicolumn{1}{c|}{99.44}           & \multicolumn{1}{c|}{99.65}            \\
			\multicolumn{1}{|c|}{}                            & \multicolumn{1}{c|}{}                            & \multicolumn{1}{c|}{Day vs. Office}            & \multicolumn{1}{c|}{77.04}           & \multicolumn{1}{c|}{90.78}           & \multicolumn{1}{c|}{94.67}            \\
			\multicolumn{1}{|c|}{}                            & \multicolumn{1}{c|}{}                            & \multicolumn{1}{c|}{Office vs. Day}            & \multicolumn{1}{c|}{84.93}           & \multicolumn{1}{c|}{94.58}           & \multicolumn{1}{c|}{97.26}            \\ \hline
			\multicolumn{1}{|c|}{\multirow{4}{*}{AdaCos}}     & \multicolumn{1}{c|}{\multirow{4}{*}{99.75}}      & \multicolumn{1}{c|}{Office vs. Office}         & \multicolumn{1}{c|}{89.71}           & \multicolumn{1}{c|}{97.22}           & \multicolumn{1}{c|}{98.66}            \\
			\multicolumn{1}{|c|}{}                            & \multicolumn{1}{c|}{}                            & \multicolumn{1}{c|}{Day. vs. Day}              & \multicolumn{1}{c|}{95.55}           & \multicolumn{1}{c|}{99.03}           & \multicolumn{1}{c|}{99.57}            \\
			\multicolumn{1}{|c|}{}                            & \multicolumn{1}{c|}{}                            & \multicolumn{1}{c|}{Day vs. Office}            & \multicolumn{1}{c|}{69.27}           & \multicolumn{1}{c|}{88.67}           & \multicolumn{1}{c|}{93.80}            \\
			\multicolumn{1}{|c|}{}                            & \multicolumn{1}{c|}{}                            & \multicolumn{1}{c|}{Office vs. Day}            & \multicolumn{1}{c|}{77.73}           & \multicolumn{1}{c|}{90.84}           & \multicolumn{1}{c|}{94.56}            \\  \hline
			\multicolumn{1}{|c|}{\multirow{4}{*}{Sofmax}}     & \multicolumn{1}{c|}{\multirow{4}{*}{99.58}}      & \multicolumn{1}{c|}{Office vs. Office}         & \multicolumn{1}{c|}{96.36}           & \multicolumn{1}{c|}{98.42}           & \multicolumn{1}{c|}{98.99}            \\
			\multicolumn{1}{|c|}{}                            & \multicolumn{1}{c|}{}                            & \multicolumn{1}{c|}{Day. vs. Day}              & \multicolumn{1}{c|}{99.08}           & \multicolumn{1}{c|}{99.83}           & \multicolumn{1}{c|}{99.85}            \\
			\multicolumn{1}{|c|}{}                            & \multicolumn{1}{c|}{}                            & \multicolumn{1}{c|}{Day vs. Office}            & \multicolumn{1}{c|}{85.02}           & \multicolumn{1}{c|}{96.20}           & \multicolumn{1}{c|}{98.31}            \\
			\multicolumn{1}{|c|}{}                            & \multicolumn{1}{c|}{}                            & \multicolumn{1}{c|}{Office vs. Day}            & \multicolumn{1}{c|}{95.19}           & \multicolumn{1}{c|}{98.97}           & \multicolumn{1}{c|}{99.32}            \\ \hline
			\multicolumn{6}{|c|}{\textbf{ResNet-50  -  VGGFace2}}                                                                                                                                                                                                                       \\ \hline
			\multicolumn{1}{|c|}{\multirow{4}{*}{CosFace}}    & \multicolumn{1}{c|}{\multirow{4}{*}{97.83}}      & \multicolumn{1}{c|}{Office vs. Office}         & \multicolumn{1}{c|}{94.25}           & \multicolumn{1}{c|}{98.32}           & \multicolumn{1}{c|}{99.02}            \\
			\multicolumn{1}{|c|}{}                            & \multicolumn{1}{c|}{}                            & \multicolumn{1}{c|}{Day. vs. Day}              & \multicolumn{1}{c|}{98.82}           & \multicolumn{1}{c|}{99.79}           & \multicolumn{1}{c|}{99.83}            \\
			\multicolumn{1}{|c|}{}                            & \multicolumn{1}{c|}{}                            & \multicolumn{1}{c|}{Day vs. Office}            & \multicolumn{1}{c|}{76.39}           & \multicolumn{1}{c|}{93.57}           & \multicolumn{1}{c|}{96.96}            \\
			\multicolumn{1}{|c|}{}                            & \multicolumn{1}{c|}{}                            & \multicolumn{1}{c|}{Office vs. Day}            & \multicolumn{1}{c|}{83.40}           & \multicolumn{1}{c|}{94.78}           & \multicolumn{1}{c|}{98.39}            \\ \hline
			\multicolumn{1}{|c|}{\multirow{4}{*}{ArcFace}}    & \multicolumn{1}{c|}{\multirow{4}{*}{97.58}}      & \multicolumn{1}{c|}{Office vs. Office}         & \multicolumn{1}{c|}{94.16}           & \multicolumn{1}{c|}{98.37}           & \multicolumn{1}{c|}{99.07}            \\
			\multicolumn{1}{|c|}{}                            & \multicolumn{1}{c|}{}                            & \multicolumn{1}{c|}{Day. vs. Day}              & \multicolumn{1}{c|}{98.73}           & \multicolumn{1}{c|}{99.74}           & \multicolumn{1}{c|}{99.87}            \\
			\multicolumn{1}{|c|}{}                            & \multicolumn{1}{c|}{}                            & \multicolumn{1}{c|}{Day vs. Office}            & \multicolumn{1}{c|}{75.49}           & \multicolumn{1}{c|}{93.39}           & \multicolumn{1}{c|}{97.10}            \\
			\multicolumn{1}{|c|}{}                            & \multicolumn{1}{c|}{}                            & \multicolumn{1}{c|}{Office vs. Day}            & \multicolumn{1}{c|}{83.61}           & \multicolumn{1}{c|}{95.63}           & \multicolumn{1}{c|}{98.78}            \\ \hline
			\multicolumn{1}{|c|}{\multirow{4}{*}{SphereFace}} & \multicolumn{1}{c|}{\multirow{4}{*}{99.13}}      & \multicolumn{1}{c|}{Office vs. Office}         & \multicolumn{1}{c|}{94.25}           & \multicolumn{1}{c|}{98.39}           & \multicolumn{1}{c|}{99.07}            \\
			\multicolumn{1}{|c|}{}                            & \multicolumn{1}{c|}{}                            & \multicolumn{1}{c|}{Day. vs. Day}              & \multicolumn{1}{c|}{98.84}           & \multicolumn{1}{c|}{99.72}           & \multicolumn{1}{c|}{99.89}            \\
			\multicolumn{1}{|c|}{}                            & \multicolumn{1}{c|}{}                            & \multicolumn{1}{c|}{Day vs. Office}            & \multicolumn{1}{c|}{77.92}           & \multicolumn{1}{c|}{93.96}           & \multicolumn{1}{c|}{97.33}            \\
			\multicolumn{1}{|c|}{}                            & \multicolumn{1}{c|}{}                            & \multicolumn{1}{c|}{Office vs. Day}            & \multicolumn{1}{c|}{84.16}           & \multicolumn{1}{c|}{96.02}           & \multicolumn{1}{c|}{98.72}            \\ \hline
			\multicolumn{1}{|c|}{\multirow{4}{*}{AdaCos}}     & \multicolumn{1}{c|}{\multirow{4}{*}{99.58}}      & \multicolumn{1}{c|}{Office vs. Office}         & \multicolumn{1}{c|}{92.43}           & \multicolumn{1}{c|}{97.84}           & \multicolumn{1}{c|}{98.78}            \\
			\multicolumn{1}{|c|}{}                            & \multicolumn{1}{c|}{}                            & \multicolumn{1}{c|}{Day. vs. Day}              & \multicolumn{1}{c|}{98.26}           & \multicolumn{1}{c|}{99.78}           & \multicolumn{1}{c|}{99.85}            \\
			\multicolumn{1}{|c|}{}                            & \multicolumn{1}{c|}{}                            & \multicolumn{1}{c|}{Day vs. Office}            & \multicolumn{1}{c|}{72.84}           & \multicolumn{1}{c|}{92.12}           & \multicolumn{1}{c|}{96.49}            \\
			\multicolumn{1}{|c|}{}                            & \multicolumn{1}{c|}{}                            & \multicolumn{1}{c|}{Office vs. Day}            & \multicolumn{1}{c|}{80.12}           & \multicolumn{1}{c|}{94.84}           & \multicolumn{1}{c|}{97.98}            \\ \hline
			\multicolumn{1}{|c|}{\multirow{4}{*}{Sofmax}}     & \multicolumn{1}{c|}{\multirow{4}{*}{99.63}}      & \multicolumn{1}{c|}{Office vs. Office}         & \multicolumn{1}{c|}{96.33}           & \multicolumn{1}{c|}{98.47}           & \multicolumn{1}{c|}{99.00}            \\
			\multicolumn{1}{|c|}{}                            & \multicolumn{1}{c|}{}                            & \multicolumn{1}{c|}{Day. vs. Day}              & \multicolumn{1}{c|}{99.36}           & \multicolumn{1}{c|}{99.81}           & \multicolumn{1}{c|}{99.85}            \\
			\multicolumn{1}{|c|}{}                            & \multicolumn{1}{c|}{}                            & \multicolumn{1}{c|}{Day vs. Office}            & \multicolumn{1}{c|}{84.25}           & \multicolumn{1}{c|}{95.61}           & \multicolumn{1}{c|}{98.06}            \\
			\multicolumn{1}{|c|}{}                            & \multicolumn{1}{c|}{}                            & \multicolumn{1}{c|}{Office vs. Day}            & \multicolumn{1}{c|}{94.45}           & \multicolumn{1}{c|}{98.99}           & \multicolumn{1}{c|}{99.63}            \\ \hline
			\multicolumn{1}{l}{}                              &                                                  
		\end{tabular}
		\label{finetun_BWC}
	\end{table*}

	\begin{table*}[h]
		\centering
		\caption{Rank-1, Rank-5 and Rank-10 accuracy values of the SENet-50 trained on MS1Mv2~\cite{deng_arcface:_2019} and VGGFace2~\cite{cao2018vggface2} datasets, fine-tuned and evaluated on non-overlapping subjects from BWCFace dataset.}
		\begin{tabular}{cclccc}
			\hline
			\multicolumn{1}{|c|}{\textbf{Loss Function}}      & \multicolumn{1}{c|}{\textbf{Validation Accuracy [\%]}} & \multicolumn{1}{c|}{\textbf{Light. Condition}} & \multicolumn{1}{c|}{\textbf{Rank-1 [\%]}} & \multicolumn{1}{c|}{\textbf{Rank-5 [\%]}} & \multicolumn{1}{c|}{\textbf{Rank-10 [\%]}} \\ \hline
			\multicolumn{6}{|c|}{\textbf{SENet-50  -  MS1Mv2}}                                                                                                                                                                                                                            \\ \hline
			\multicolumn{1}{|c|}{\multirow{4}{*}{CosFace}}    & \multicolumn{1}{c|}{\multirow{4}{*}{95.29}}      & \multicolumn{1}{l|}{Office vs. Office}         & \multicolumn{1}{l|}{93.431}          & \multicolumn{1}{l|}{97.902}          & \multicolumn{1}{l|}{98.755}           \\
			\multicolumn{1}{|c|}{}                            & \multicolumn{1}{c|}{}                            & \multicolumn{1}{l|}{Day. vs. Day}              & \multicolumn{1}{l|}{97.665}          & \multicolumn{1}{l|}{99.570}          & \multicolumn{1}{l|}{99.851}           \\
			\multicolumn{1}{|c|}{}                            & \multicolumn{1}{c|}{}                            & \multicolumn{1}{l|}{Day vs. Office}            & \multicolumn{1}{l|}{78.549}          & \multicolumn{1}{l|}{93.373}          & \multicolumn{1}{l|}{96.804}           \\
			\multicolumn{1}{|c|}{}                            & \multicolumn{1}{c|}{}                            & \multicolumn{1}{l|}{Office vs. Day}            & \multicolumn{1}{l|}{85.770}          & \multicolumn{1}{l|}{94.991}          & \multicolumn{1}{l|}{97.748}           \\ \hline
			\multicolumn{1}{|c|}{\multirow{4}{*}{ArcFace}}    & \multicolumn{1}{c|}{\multirow{4}{*}{94.63}}      & \multicolumn{1}{l|}{Office vs. Office}         & \multicolumn{1}{l|}{92.754}          & \multicolumn{1}{l|}{97.588}          & \multicolumn{1}{l|}{98.598}           \\
			\multicolumn{1}{|c|}{}                            & \multicolumn{1}{c|}{}                            & \multicolumn{1}{l|}{Day. vs. Day}              & \multicolumn{1}{l|}{97.422}          & \multicolumn{1}{l|}{99.514}          & \multicolumn{1}{l|}{99.813}           \\
			\multicolumn{1}{|c|}{}                            & \multicolumn{1}{c|}{}                            & \multicolumn{1}{l|}{Day vs. Office}            & \multicolumn{1}{l|}{77.627}          & \multicolumn{1}{l|}{93.510}          & \multicolumn{1}{l|}{97.137}           \\
			\multicolumn{1}{|c|}{}                            & \multicolumn{1}{c|}{}                            & \multicolumn{1}{l|}{Office vs. Day}            & \multicolumn{1}{l|}{84.586}          & \multicolumn{1}{l|}{94.448}          & \multicolumn{1}{l|}{97.554}           \\ \hline
			\multicolumn{1}{|c|}{\multirow{4}{*}{SphereFace}} & \multicolumn{1}{c|}{\multirow{4}{*}{97.54}}      & \multicolumn{1}{l|}{Office vs. Office}         & \multicolumn{1}{l|}{93.362}          & \multicolumn{1}{l|}{97.990}          & \multicolumn{1}{l|}{98.774}           \\
			\multicolumn{1}{|c|}{}                            & \multicolumn{1}{c|}{}                            & \multicolumn{1}{l|}{Day. vs. Day}              & \multicolumn{1}{l|}{97.646}          & \multicolumn{1}{l|}{99.514}          & \multicolumn{1}{l|}{99.795}           \\
			\multicolumn{1}{|c|}{}                            & \multicolumn{1}{c|}{}                            & \multicolumn{1}{l|}{Day vs. Office}            & \multicolumn{1}{l|}{78.529}          & \multicolumn{1}{l|}{93.490}          & \multicolumn{1}{l|}{97.000}           \\
			\multicolumn{1}{|c|}{}                            & \multicolumn{1}{c|}{}                            & \multicolumn{1}{l|}{Office vs. Day}            & \multicolumn{1}{l|}{84.974}          & \multicolumn{1}{l|}{94.040}          & \multicolumn{1}{l|}{97.204}           \\ \hline
			\multicolumn{1}{|c|}{\multirow{4}{*}{AdaCos}}     & \multicolumn{1}{c|}{\multirow{4}{*}{98.92}}      & \multicolumn{1}{l|}{Office vs. Office}         & \multicolumn{1}{l|}{90.666}          & \multicolumn{1}{l|}{97.039}          & \multicolumn{1}{l|}{98.157}           \\
			\multicolumn{1}{|c|}{}                            & \multicolumn{1}{c|}{}                            & \multicolumn{1}{l|}{Day. vs. Day}              & \multicolumn{1}{l|}{96.507}          & \multicolumn{1}{l|}{99.477}          & \multicolumn{1}{l|}{99.720}           \\
			\multicolumn{1}{|c|}{}                            & \multicolumn{1}{c|}{}                            & \multicolumn{1}{l|}{Day vs. Office}            & \multicolumn{1}{l|}{73.333}          & \multicolumn{1}{l|}{91.549}          & \multicolumn{1}{l|}{96.000}           \\
			\multicolumn{1}{|c|}{}                            & \multicolumn{1}{c|}{}                            & \multicolumn{1}{l|}{Office vs. Day}            & \multicolumn{1}{l|}{81.208}          & \multicolumn{1}{l|}{92.972}          & \multicolumn{1}{l|}{96.525}           \\ \hline
			\multicolumn{1}{|c|}{\multirow{4}{*}{Sofmax}}     & \multicolumn{1}{c|}{\multirow{4}{*}{99.21}}      & \multicolumn{1}{l|}{Office vs. Office}         & \multicolumn{1}{l|}{95.000}          & \multicolumn{1}{l|}{98.127}          & \multicolumn{1}{l|}{98.755}           \\
			\multicolumn{1}{|c|}{}                            & \multicolumn{1}{c|}{}                            & \multicolumn{1}{l|}{Day. vs. Day}              & \multicolumn{1}{l|}{99.197}          & \multicolumn{1}{l|}{99.738}          & \multicolumn{1}{l|}{99.813}           \\
			\multicolumn{1}{|c|}{}                            & \multicolumn{1}{c|}{}                            & \multicolumn{1}{l|}{Day vs. Office}            & \multicolumn{1}{l|}{87.392}          & \multicolumn{1}{l|}{96.059}          & \multicolumn{1}{l|}{98.039}           \\
			\multicolumn{1}{|c|}{}                            & \multicolumn{1}{c|}{}                            & \multicolumn{1}{l|}{Office vs. Day}            & \multicolumn{1}{l|}{94.545}          & \multicolumn{1}{l|}{98.796}          & \multicolumn{1}{l|}{99.340}           \\ \hline
			\multicolumn{6}{|c|}{\textbf{SENet-50  -  VGGFace2}}                                                                                                                                                                                                                        \\ \hline
			\multicolumn{1}{|c|}{\multirow{4}{*}{CosFace}}    & \multicolumn{1}{c|}{\multirow{4}{*}{96.96}}      & \multicolumn{1}{l|}{Office vs. Office}         & \multicolumn{1}{l|}{93.990}          & \multicolumn{1}{l|}{98.402}          & \multicolumn{1}{l|}{99.176}           \\
			\multicolumn{1}{|c|}{}                            & \multicolumn{1}{c|}{}                            & \multicolumn{1}{l|}{Day. vs. Day}              & \multicolumn{1}{l|}{97.740}          & \multicolumn{1}{l|}{99.514}          & \multicolumn{1}{l|}{99.738}           \\
			\multicolumn{1}{|c|}{}                            & \multicolumn{1}{c|}{}                            & \multicolumn{1}{l|}{Day vs. Office}            & \multicolumn{1}{l|}{79.353}          & \multicolumn{1}{l|}{94.235}          & \multicolumn{1}{l|}{97.569}           \\
			\multicolumn{1}{|c|}{}                            & \multicolumn{1}{c|}{}                            & \multicolumn{1}{l|}{Office vs. Day}            & \multicolumn{1}{l|}{86.158}          & \multicolumn{1}{l|}{96.738}          & \multicolumn{1}{l|}{98.777}           \\ \hline
			\multicolumn{1}{|c|}{\multirow{4}{*}{ArcFace}}    & \multicolumn{1}{c|}{\multirow{4}{*}{97.08}}      & \multicolumn{1}{l|}{Office vs. Office}         & \multicolumn{1}{l|}{93.372}          & \multicolumn{1}{l|}{98.039}          & \multicolumn{1}{l|}{99.049}           \\
			\multicolumn{1}{|c|}{}                            & \multicolumn{1}{c|}{}                            & \multicolumn{1}{l|}{Day. vs. Day}              & \multicolumn{1}{l|}{97.777}          & \multicolumn{1}{l|}{99.570}          & \multicolumn{1}{l|}{99.776}           \\
			\multicolumn{1}{|c|}{}                            & \multicolumn{1}{c|}{}                            & \multicolumn{1}{l|}{Day vs. Office}            & \multicolumn{1}{l|}{78.725}          & \multicolumn{1}{l|}{93.588}          & \multicolumn{1}{l|}{97.471}           \\
			\multicolumn{1}{|c|}{}                            & \multicolumn{1}{c|}{}                            & \multicolumn{1}{l|}{Office vs. Day}            & \multicolumn{1}{l|}{85.673}          & \multicolumn{1}{l|}{96.855}          & \multicolumn{1}{l|}{98.758}           \\ \hline
			\multicolumn{1}{|c|}{\multirow{4}{*}{SphereFace}} & \multicolumn{1}{c|}{\multirow{4}{*}{98.69}}      & \multicolumn{1}{l|}{Office vs. Office}         & \multicolumn{1}{l|}{94.019}          & \multicolumn{1}{l|}{98.294}          & \multicolumn{1}{l|}{99.196}           \\
			\multicolumn{1}{|c|}{}                            & \multicolumn{1}{c|}{}                            & \multicolumn{1}{l|}{Day. vs. Day}              & \multicolumn{1}{l|}{97.553}          & \multicolumn{1}{l|}{99.477}          & \multicolumn{1}{l|}{99.776}           \\
			\multicolumn{1}{|c|}{}                            & \multicolumn{1}{c|}{}                            & \multicolumn{1}{l|}{Day vs. Office}            & \multicolumn{1}{l|}{78.549}          & \multicolumn{1}{l|}{93.373}          & \multicolumn{1}{l|}{97.235}           \\
			\multicolumn{1}{|c|}{}                            & \multicolumn{1}{c|}{}                            & \multicolumn{1}{l|}{Office vs. Day}            & \multicolumn{1}{l|}{85.867}          & \multicolumn{1}{l|}{96.331}          & \multicolumn{1}{l|}{98.544}           \\ \hline
			\multicolumn{1}{|c|}{\multirow{4}{*}{AdaCos}}     & \multicolumn{1}{c|}{\multirow{4}{*}{99.50}}      & \multicolumn{1}{l|}{Office vs. Office}         & \multicolumn{1}{l|}{91.107}          & \multicolumn{1}{l|}{97.598}          & \multicolumn{1}{l|}{98.774}           \\
			\multicolumn{1}{|c|}{}                            & \multicolumn{1}{c|}{}                            & \multicolumn{1}{l|}{Day. vs. Day}              & \multicolumn{1}{l|}{96.544}          & \multicolumn{1}{l|}{99.346}          & \multicolumn{1}{l|}{99.664}           \\
			\multicolumn{1}{|c|}{}                            & \multicolumn{1}{c|}{}                            & \multicolumn{1}{l|}{Day vs. Office}            & \multicolumn{1}{l|}{72.863}          & \multicolumn{1}{l|}{92.216}          & \multicolumn{1}{l|}{96.510}           \\
			\multicolumn{1}{|c|}{}                            & \multicolumn{1}{c|}{}                            & \multicolumn{1}{l|}{Office vs. Day}            & \multicolumn{1}{l|}{83.479}          & \multicolumn{1}{l|}{96.059}          & \multicolumn{1}{l|}{98.466}           \\ \hline
			\multicolumn{1}{|c|}{\multirow{4}{*}{Sofmax}}     & \multicolumn{1}{c|}{\multirow{4}{*}{99.85}}      & \multicolumn{1}{l|}{Office vs. Office}         & \multicolumn{1}{l|}{96.088}          & \multicolumn{1}{l|}{98.686}          & \multicolumn{1}{l|}{99.245}           \\
			\multicolumn{1}{|c|}{}                            & \multicolumn{1}{c|}{}                            & \multicolumn{1}{l|}{Day. vs. Day}              & \multicolumn{1}{l|}{98.767}          & \multicolumn{1}{l|}{99.738}          & \multicolumn{1}{l|}{99.813}           \\
			\multicolumn{1}{|c|}{}                            & \multicolumn{1}{c|}{}                            & \multicolumn{1}{l|}{Day vs. Office}            & \multicolumn{1}{l|}{84.333}          & \multicolumn{1}{l|}{96.176}          & \multicolumn{1}{l|}{98.412}           \\
			\multicolumn{1}{|c|}{}                            & \multicolumn{1}{c|}{}                            & \multicolumn{1}{l|}{Office vs. Day}            & \multicolumn{1}{l|}{91.322}          & \multicolumn{1}{l|}{98.156}          & \multicolumn{1}{l|}{99.495}           \\ \hline
			
		\end{tabular}
		
	\end{table*}

\section{Conclusion}
This is the first study evaluating open-set face recognition on our BWCFace data captured using the body-worn camera. To this aim, the latest deep learning models with different loss functions are evaluated on our dataset. Experimental results suggest a maximum of $33.89\%$ Rank-1 accuracy obtained. The reason could be datasets such as VGGFace2 and MS1Mv2 scrapped from the web do not contain images captured from Body-worn cameras. Therefore, these datasets may not be representative of the samples acquired using a body-worn camera.  However, on fine-tuning the models using a small subset of subjects captured using the body-worn camera, hallmark Rank-1 accuracy of $99\%$ was obtained. Further, higher accuracy rates were also obtained on data samples captured from a different model of the body-worn camera from the UBHSD dataset~\cite{Al-Obaydy11}. No significant differences were observed across the loss functions. As a part of future work, lightweight CNN models such as lightCNN~\cite{8353856} will be evaluated. Cross-sensor evaluation will be performed using data captured by mobile devices~\cite{selfie_bio}. Note that the study on the bias of the face recognition technology across demographic variations is a different research topic that is beyond the scope of this current study. 


	\section*{Acknowledgment}
	This work is supported in part by Award for Research/Creative projects by Wichita State University.

	
	
	%

\end{document}